\documentclass[10pt,twocolumn,letterpaper]{article}

\usepackage{cvpr}              
\usepackage{booktabs}
\usepackage{graphicx, amsmath, amssymb, caption, subcaption, multirow, overpic, textpos, pifont, adjustbox}
\usepackage{colortbl}
\usepackage[pagebackref,breaklinks,colorlinks]{hyperref}

\newlength\savewidth
\newcommand{\tablestyle}[2]{\setlength{\tabcolsep}{#1}\renewcommand{\arraystretch}{#2}\centering\footnotesize}

\renewcommand{\paragraph}[1]{\vspace{1.25mm}\noindent\textbf{#1}}
\newcommand{\app}{\raise.17ex\hbox{$\scriptstyle\sim$}}

\newcommand{\sam}{SAM\xspace}
\newcommand{\fig}[1]{Fig.~\ref{#1}}

\newcommand{\demph}[1]{\textcolor{demphcolor}{#1}}
\newcommand{\mypm}[1]{{\scriptsize{{\demph{{\kern.4ex$\pm$\kern.1ex#1}}}}}}

\usepackage[capitalize]{cleveref}
\crefname{section}{Sec.}{Secs.}
\Crefname{section}{Section}{Sections}
\Crefname{table}{Table}{Tables}
\crefname{table}{Tab.}{Tabs.}

\begin{document}

\title{\vspace{-3mm} Fast Segment Anything \vspace{-6mm}}
\author{Xu Zhao $^{1,3}$ \quad  Wenchao Ding $^{1,2}$  \quad Yongqi An $^{1,2}$ \quad Yinglong Du $^{1,2}$ \\
 \quad Tao Yu $^{1,2}$ \quad Min Li $^{1,2}$ \quad Ming Tang $^{1,2}$\quad Jinqiao Wang $^{1,2,3,4}$\\
   Institute of Automation, Chinese Academy of Sciences, Beijing, China$^{1}$ \\ 
  School of Artificial Intelligence, University of Chinese Academy of Sciences, Beijing, China$^{2}$\\
   Objecteye  Inc., Beijing, China$^{3}$\\
   Wuhan AI Research, Wuhan, China$^{4}$\\
   \\
{\tt\small \{xu.zhao,yongqi.an,tangm,jqwang\}@nlpr.ia.ac.cn}\\
{\tt\small  \{dingwenchao2021,duyinglong2022,yutao2022,limin2021\}@ia.ac.cn
}
}

\maketitle

\begin{abstract}\vspace{-3mm}
   The recently proposed segment anything model (SAM) has made a significant influence in many computer vision tasks. It is becoming a foundation step for many high-level tasks, like \textcolor{black}{image segmentation, image caption, and image editing}. However, its huge computation costs prevent it from wider applications in industry scenarios. The computation mainly comes from the Transformer architecture at high-resolution inputs. In this paper, we propose a speed-up alternative method for this fundamental task with comparable performance. By reformulating the task as segments-generation and prompting, we find that a regular CNN detector with an instance segmentation branch can also accomplish this task well. Specifically, we convert this task to the well-studied instance segmentation task and directly train the existing instance segmentation method using only 1/50 of the SA-1B dataset published by SAM authors. With our method, we achieve a comparable performance with the SAM method at \textbf{50$\times$} higher run-time speed. We give sufficient experimental results to demonstrate its effectiveness.  The codes and demos will be released at \url{https://github.com/CASIA-IVA-Lab/FastSAM}.
\vspace{-3mm}
\end{abstract}

\section{Introduction}\label{sec:intro}

\begin{figure}
    \centering
    \includegraphics[width=\columnwidth]{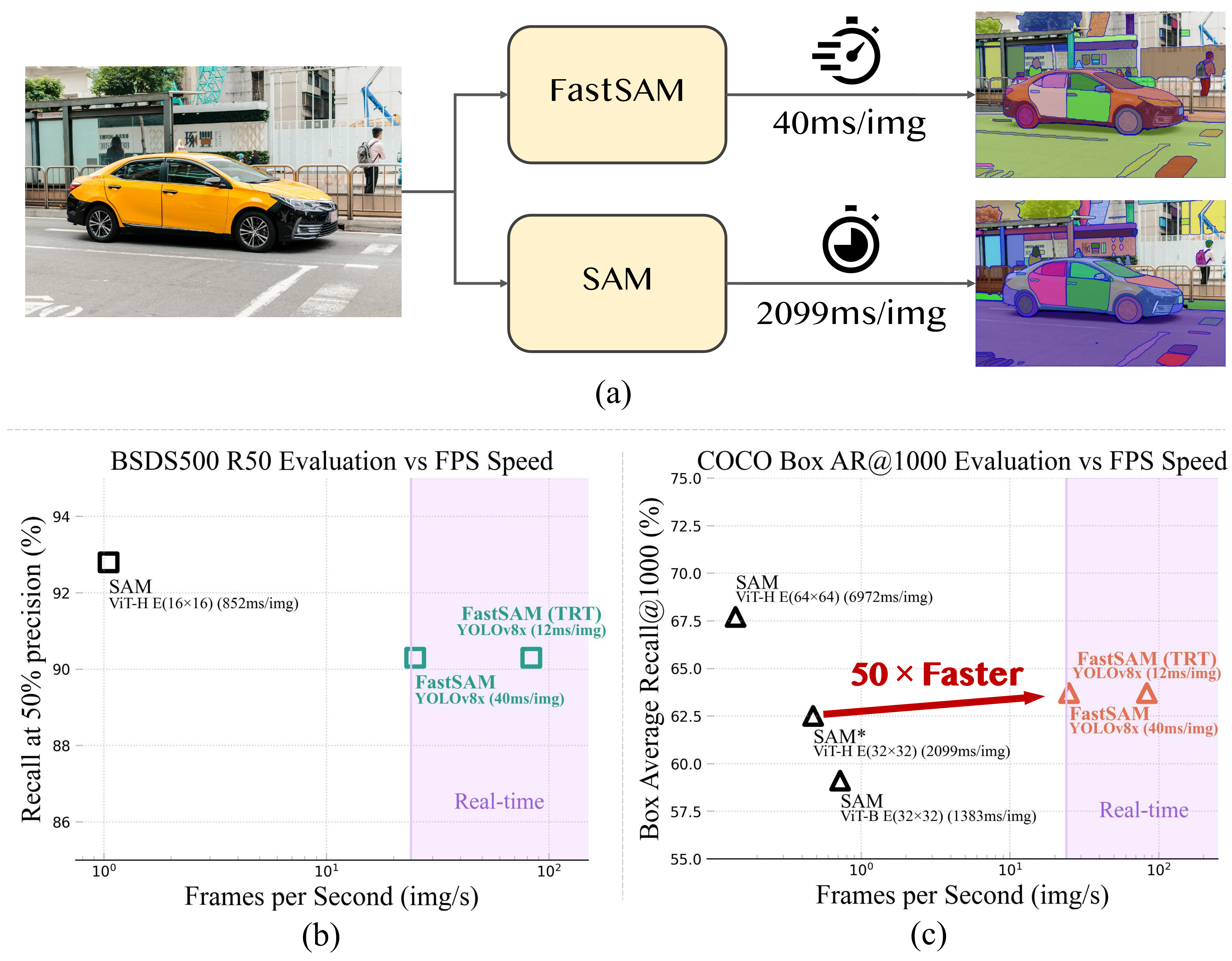}
    \caption{Comparative analysis of FastSAM and SAM. (a) Speed comparison between FastSAM and SAM on a single NVIDIA GeForce RTX 3090. (b) Comparison on the BSDS500 dataset \cite{martin2001database,arbelaez2010contour} for edge detection. (c) Box AR@1000 evaluation of FastSAM and SAM on the COCO dataset \cite{lin2014microsoft} for the object proposal. Both SAM and FastSAM are tested using PyTorch for inference, except FastSAM(TRT) uses TensorRT for inference.}
    \label{fig:head}
\end{figure}

Recently, the \emph{Segment Anything Model} (SAM)~\cite{kirillov2023segment} is proposed. It is regarded as a milestone vision foundation model. It can segment any object within the image guided by various possible user interaction prompts. SAM leverages a Transformer model trained on the extensive SA-1B dataset, which gives it the ability to deftly handle a wide range of scenes and objects. SAM opens the door to an exciting new task known as \emph{Segment Anything}. This task, due to its generalizability and potentiality, has all the makings of becoming a cornerstone for a broad spectrum of future vision tasks.

However, despite these advancements and the promising results shown by SAM and subsequent models in handling the segment anything task, its practical applications are still challenging. The glaring issue is the substantial computational resource requirements associated with Transformer (ViT) models, the main part of SAM's architecture. When compared with their convolutional counterparts, ViTs stand out for their heavy computation resources demands, which presents a hurdle to their practical deployment, especially in real-time applications. This limitation consequently hinders the progress and potential of the {segment anything} task.

Motivated by the high demand from the industrial applications for the segment anything model, in this paper we design a real-time solution for the {segment anything} task, FastSAM. We decouple the segment anything task into two sequential stages which are all-instance segmentation and prompt-guided selection. The first stage hinges on the implementation of a Convolutional Neural Network (CNN)-based detector. It produces the segmentation masks of all instances in the image. Then in the second stage, it output the the region-of-interest corresponding the prompt. By leveraging the computational efficiency of CNNs, we demonstrate that a real-time segment of anything model is achievable without much compromising on performance quality. We hope the proposed method would facilitate the industrial applications of the foundational task of segmenting anything.

Our proposed FastSAM is based on YOLOv8-seg~\cite{yolov8_ultralytics1},
an object detector equipped with the instance segmentation branch, which utilizes the YOLACT~\cite{bolya2019yolact} method. We also adopt the extensive SA-1B dataset published by SAM. 
By directly training this CNN detector on only 2\% (1/50) of the SA-1B dataset, it achieves comparable performance to SAM, but with drastically reduced computational and resource demands,
thereby enabling real-time application. We also apply it to multiple downstream segmentation tasks to show its generalization performance. 
On the object proposal task on MS COCO~\cite{gupta2019lvis}, we achieve 63.7 at AR1000, which is 1.2 points higher than SAM with 32$\times$ 32 point-prompt inputs, but running 50 times faster on a single NVIDIA RTX 3090.

The real-time segment anything model is valuable for industrial applications. It can be applied to many scenarios. The proposed approach not only provides a new, practical solution for a large number of vision tasks but also does so at a really high speed, tens or hundreds of times faster than current methods. 

It also offers new views for the large model architecture for the general vision tasks. We think for specific tasks, specific models still take advantage to get a better efficiency-accuracy trade-off.

Then, in the sense of model compression, our approach demonstrates the feasibility of a path that can significantly reduce the computational effort by introducing an artificial prior to the structure.

Our contributions can be summarized as follow:
\begin{itemize}
    \setlength{\itemsep}{0pt}
    \setlength{\topsep}{0pt}
    \setlength{\parsep}{0pt}
    \item A novel, real-time CNN-based solution for the \emph{Segment Anything} task is introduced, which significantly reduces computational demands while maintaining competitive performance.
    \item This work presents the first study of applying a CNN detector to the segment anything task, offering insights into the potential of lightweight CNN models in complex vision tasks.
    \item A comparative evaluation between the proposed method and SAM on multiple benchmarks provides insights into the strengths and weaknesses of the approach in the segment anything domain.
\end{itemize}


\section{Preliminary}\label{sec:background}

In this section, we give a review of the segment anything model and a clear definition of the segment anything task. 

\paragraph{Segment Anything Model.} In the evolving field of image segmentation, the Segment Anything Model (SAM)~\cite{kirillov2023segment} is a significant innovation due to its proposed training methodology and performance on large-scale visual datasets. SAM provides a high-precision, class-agnostic segmentation performance, exhibiting distinct capabilities in zero-shot tasks. As a foundation model, it expands the horizons of computer vision by showing not just powerful interactive segmentation methods, but also exceptional adaptability across a variety of segmentation tasks. SAM is a striking example of the potential of foundation models for open-world image understanding. However, while the model's performance is satisfying, it is worth noting that SAM faces a significant limitation – the lack of real-time processing capability. This restricts its wide application in scenarios where immediate segmentation results are critical. 

\paragraph{Segment Anything Task.} The \emph{Segment Anything} task is defined as a process whereby an effective segmentation mask is produced given any form of the prompt. These prompts range from foreground/background point sets, rough boxes or masks, free-form text, or any information that indicates the content to be segmented within an image. We have discovered that the segment anything task can be effectively broken down into two stages in the majority of practical applications. The first stage involves detecting and segmenting all objects in the image, like a panoptic segmentation~\cite{kirillov2019panoptic} process. The second stage depends on the provided prompts to separate the specific object(s) of interest from the segmented panorama. The decoupling of this task significantly reduces its complexity, thus providing the possibility to propose a real-time segment of anything model. 
\begin{figure*}[htb]
\centering
\includegraphics[width=0.95\linewidth]{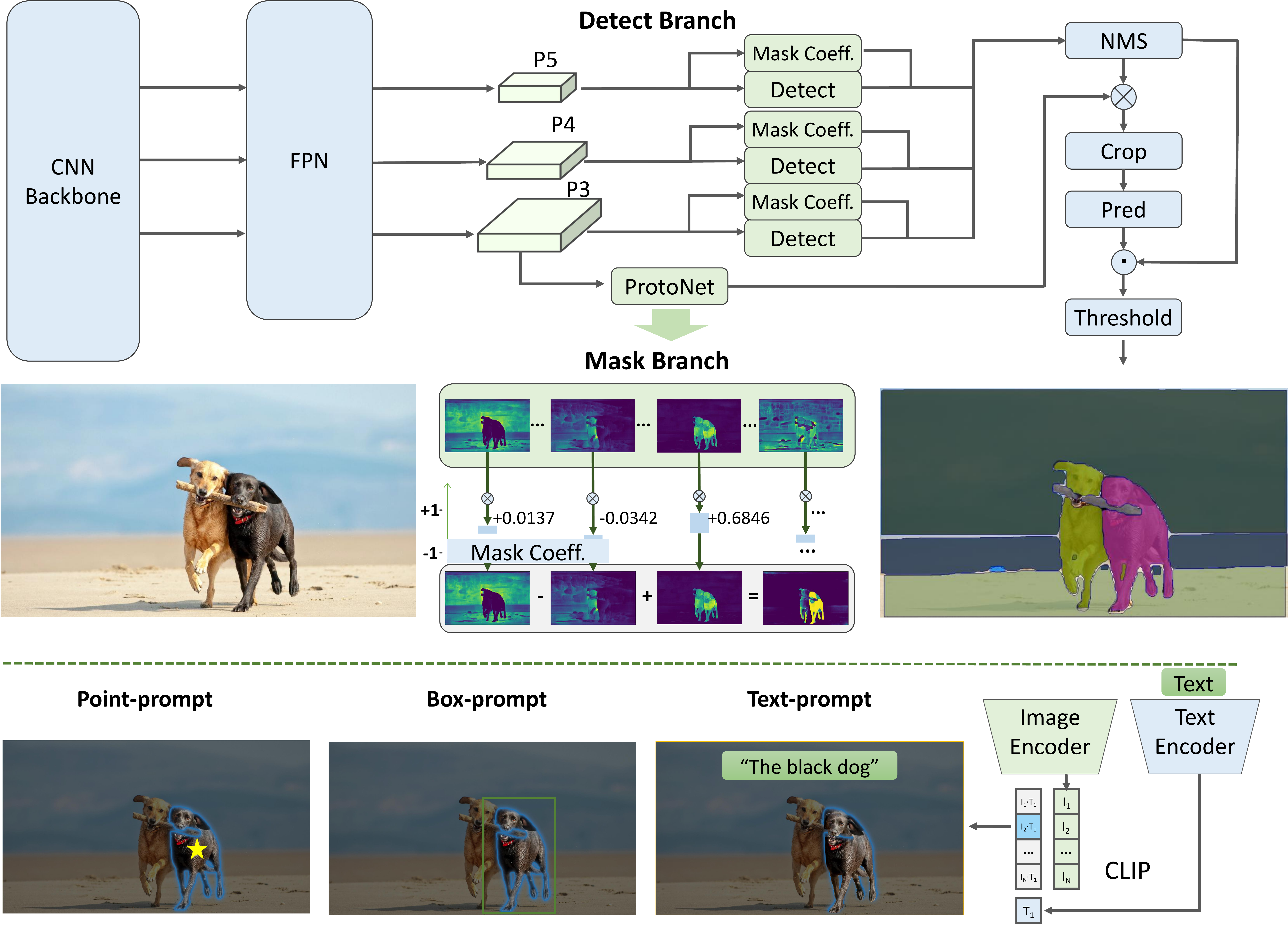} 
\caption{The framework of FastSAM. It contains two stages: All-instance Segmentation (AIS) and Prompt-guided Selection (PGS). We use YOLOv8-seg~\cite{yolov8_ultralytics1} to segment all objects or regions in an image. Then we use various prompts to identify the specific object(s) of interest. It mainly involves the utilization of point prompts, box prompts, and text prompt. The text prompt is based on CLIP~\cite{radford2021learning}.}
\label{fig:overview}
\end{figure*}
\section{Methodology}\label{sec:method}

\subsection{Overview}\label{sec:method:overview}

\fig{fig:overview} gives the overview of the proposed method, FastSAM. The method consists of two stages, i.e. the All-instance Segmentation and the Prompt-guided Selection. The former stage is a basis and the second stage is essentially the task-oriented post-processing. Different from the end-to-end transformers~\cite{cheng2021per,kirillov2023segment,carion2020end}, the overall method introduces many human priors which match the vision segmentation tasks, like the local connections of convolutions and the receptive-field-relevant object assigning strategies. This makes it tailored for the vision segmentation task and can converge faster on a smaller number of parameters.


\subsection{All-instance Segmentation}\label{sec:method:AIS}

\paragraph{Model Architecture.} The architecture of YOLOv8~\cite{yolov8_ultralytics1} develops from its predecessor, YOLOv5~\cite{yolov5}, integrating key design aspects from recent algorithms such as YOLOX~\cite{ge2021yolox}, YOLOv6~\cite{li2023yolov6}, and YOLOv7~\cite{wang2022yolov7}. YOLOv8's backbone network and neck module substitute YOLOv5's C3 module with the C2f module. The updated \emph{Head} module embraces a decoupled structure, separating classification and detection heads, and shifts from Anchor-Based to Anchor-Free.

\paragraph{Instance Segmentation.} YOLOv8-seg applies YOLACT~\cite{bolya2019yolact} principles for instance segmentation. It begins with feature extraction from an image via a backbone network and the Feature Pyramid Network (FPN)~\cite{lin2017feature}, integrating diverse size features. The output consists of the detection and segmentation branches. 

The detection branch outputs category and bounding box, while the segmentation branch outputs $k$ prototypes (defaulted to 32 in FastSAM) along with k mask coefficients. The segmentation and detection tasks are computed in parallel. The segmentation branch inputs a high-resolution feature map, preserves spatial details, and also contains semantic information. This map is processed through a convolution layer, upscaled, and then passed through two more convolution layers to output the masks. The mask coefficients, similar to the detection head's classification branch, range between -1 and 1. The instance segmentation result is obtained by multiplying the mask coefficients with the prototypes and then summing them up.

YOLOv8 can be used in a variety of object detection tasks. With the instance segmentation branch, YOLOv8-Seg is born suitable for the segment anything task, which aims to accurately detect and segment every object or region in an image, regardless of the object category. The prototypes and mask coefficients provide a lot of extensibility for prompt guidance. As a simple example, a simple prompt encoder and decoder structure is additionally trained, with various prompts and image feature embeddings as input and mask coefficients as output. In FastSAM, we directly use the YOLOv8-seg method for the all-instance segmentation stage. The more artificial design might bring additional improvements, but we regard it as out of the scope of this work and leave it for future study.

\begin{table*}[t]
\centering
\tablestyle{2.8pt}{1.1}
\footnotesize
\begin{tabular}{c|c|p{1cm}<{\centering}p{1cm}<{\centering}p{1cm}<{\centering}p{1.5cm}<{\centering}p{1.5cm}<{\centering}p{1.5cm}<{\centering}}
\toprule
& & \multicolumn{6}{c}{Running Speed under Different Point Prompt Numbers (ms)} \\

method & params & 1 & 10 & 100 & E(16$\times$16) & E(32$\times$32*)  & E(64$\times$64)   \\
\hline
 
\sam-H ~\cite{SAM} &0.6G & 446 & 464  &  627 & 852 & 2099 & 6972   \\
\sam-B ~\cite{SAM} &136M & 110 & 125  &  230 & 432 &1383 & 5417 \\
\hline
FastSAM (Ours) & 68M &  \multicolumn{6}{c}{40}  \\
\bottomrule
\end{tabular}
\vspace{-2mm}
\caption{Running Speed (ms/image) of SAM and FastSAM under different point prompt numbers. E: Everything Mode of SAM. *: 32$\times$32 is the default setting of SAM for many tasks\protect\footnotemark.}
\label{tab:speed}
\end{table*}

\subsection{Prompt-guided Selection}\label{sec:method:PGS}

Following the successful segmentation of all objects or regions in an image using YOLOv8, the second stage of the segment anything task is to use various prompts to identify the specific object(s) of interest. It mainly involves the utilization of point prompts, box prompts, and text prompts.

\paragraph{Point prompt.} The point prompt consists of matching the selected points to the various masks obtained from the first phase. The goal is to determine the mask in which the point is located. Similar to SAM, we employ foreground/background points as the prompt in our approach. In cases where a foreground point is located in multiple masks, background points can be utilized to filter out masks that are irrelevant to the task at hand. By employing a set of foreground/background points, we are able to select multiple masks within the region of interest. These masks will be merged into a single mask to completely mark the object of interest. In addition, we utilize morphological operations to improve the performance of mask merging.

\paragraph{Box prompt.} The box prompt involves performing Intersection over Union (IoU) matching between the selected box and the bounding boxes corresponding to the various masks from the first phase. The aim is to identify the mask with the highest IoU score with the selected box and thus select the object of interest.

\paragraph{Text prompt.} In the case of text prompt, the corresponding text embeddings of the text are extracted using the CLIP~\cite{radford2021learning} model. The respective image embeddings are then determined and matched to the intrinsic features of each mask using a similarity metric. The mask with the highest similarity score to the image embeddings of the text prompt is then selected.

By carefully implementing these prompt-guided selection techniques, the FastSAM can reliably select specific objects of interest from a segmented image. The above approach provides an efficient way to accomplish the segment anything task in real-time, thus greatly enhancing the utility of the YOLOv8 model for complex image segmentation tasks. A more effective prompt-guided selection technique is left for future exploration.

\section{Experiments}\label{sec:eval}
 \begin{figure*}[t]
\vspace{0.01cm}
	\centering
	\includegraphics[width=0.9\linewidth]{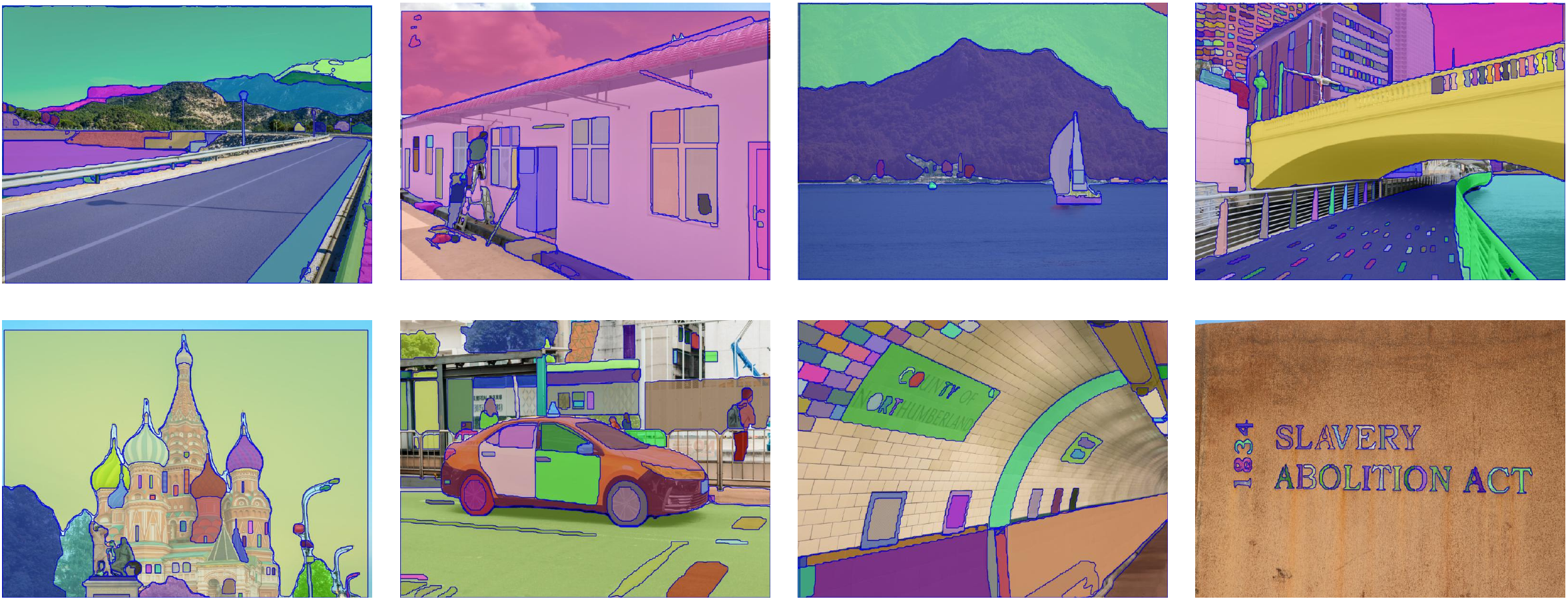}
	\vspace{-0.1cm}
	\caption{Segmentation Results of FastSAM}
	\vspace{-0.1cm}
	\label{fig:more}
\end{figure*}

In this section, we first analysis  the run-time efficiency of FastSAM. Then we experiment with four zero-shot tasks, along with applications in real-world scenarios, efficiency, and deployment. In the first part of the experiments, our goal is to test the similarity in capabilities between FastSAM and SAM. Following SAM, we also experiment with four tasks with different levels: (1) low-level: edge detection, (2) mid-level: object proposal generation, (3) high-level: instance segmentation, and finally, (4) high-level: segmenting objects with free-form text input. Our experiments also further validate FastSAM's capabilities with real-world applications and speed.

\paragraph{Implementation Details.} Unless stated otherwise, the following conditions apply: (1) FastSAM employs a YOLOv8-x~\cite{yolov8_ultralytics1} model as the main part of its architecture, with the input size of 1024; (2) FastSAM's training was carried out on
2$\%$ of the SA-1B dataset~\cite{kirillov2023segment}; (3) We train the model for 100 epochs using the default hyper-parameter settings except that the \emph{reg\_max} in the bounding box regression module is changed from 16 to 26 for predicting large instances.

\subsection{Run-time Efficiency Evaluation}\label{sec:eval:speed}
SAM uses the Transformer architecture to construct an end-to-end algorithm. The Transformer is a universal architecture that can represent many forms of mapping functions of various tasks. To segment anything, SAM learns the vision-oriented inductive bias through the learning process on large-scale data. 
On the contrary, with the human priori knowledge in structure designing, FastSAM obtains a relatively compact model. From Figure \ref{fig:more}, The FastSAM generates relatively satisfying results.

In Table \ref{tab:speed}, we report the running speed of SAM and FastSAM on a single  NVIDIA GeForce RTX 3090 GPU. It can be seen that FastSAM surpasses SAM at all prompt numbers. Moreover, the running speed of FastSAM does not change with the prompts, making it a better choice for the Everything mode.

\begin{figure*}[t]\centering
\tablestyle{1pt}{0.8}\begin{tabular}{cccc}
image & ground truth & SAM & FastSAM \\
\includegraphics[width=0.25\linewidth]{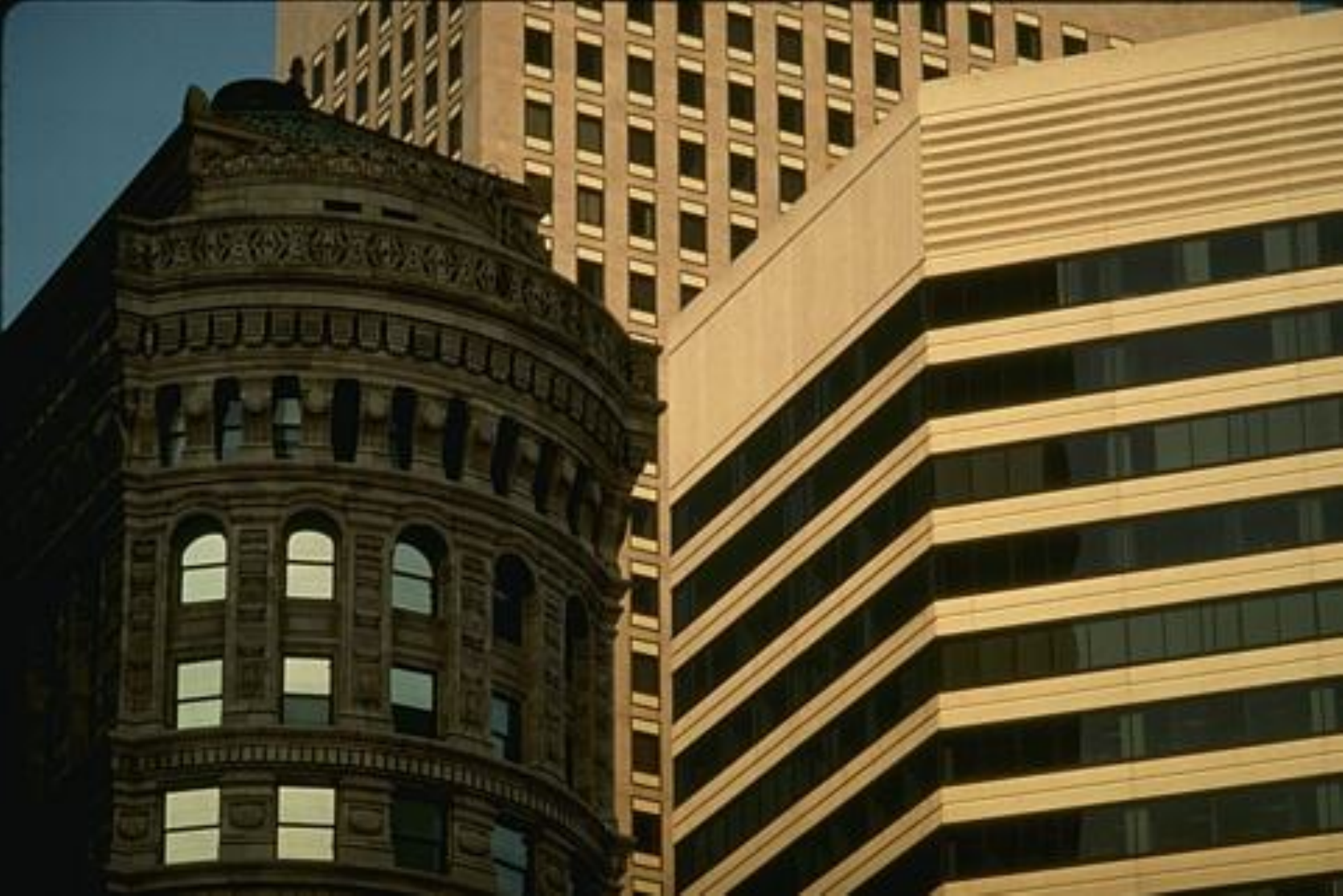} &
\includegraphics[width=0.25\linewidth]{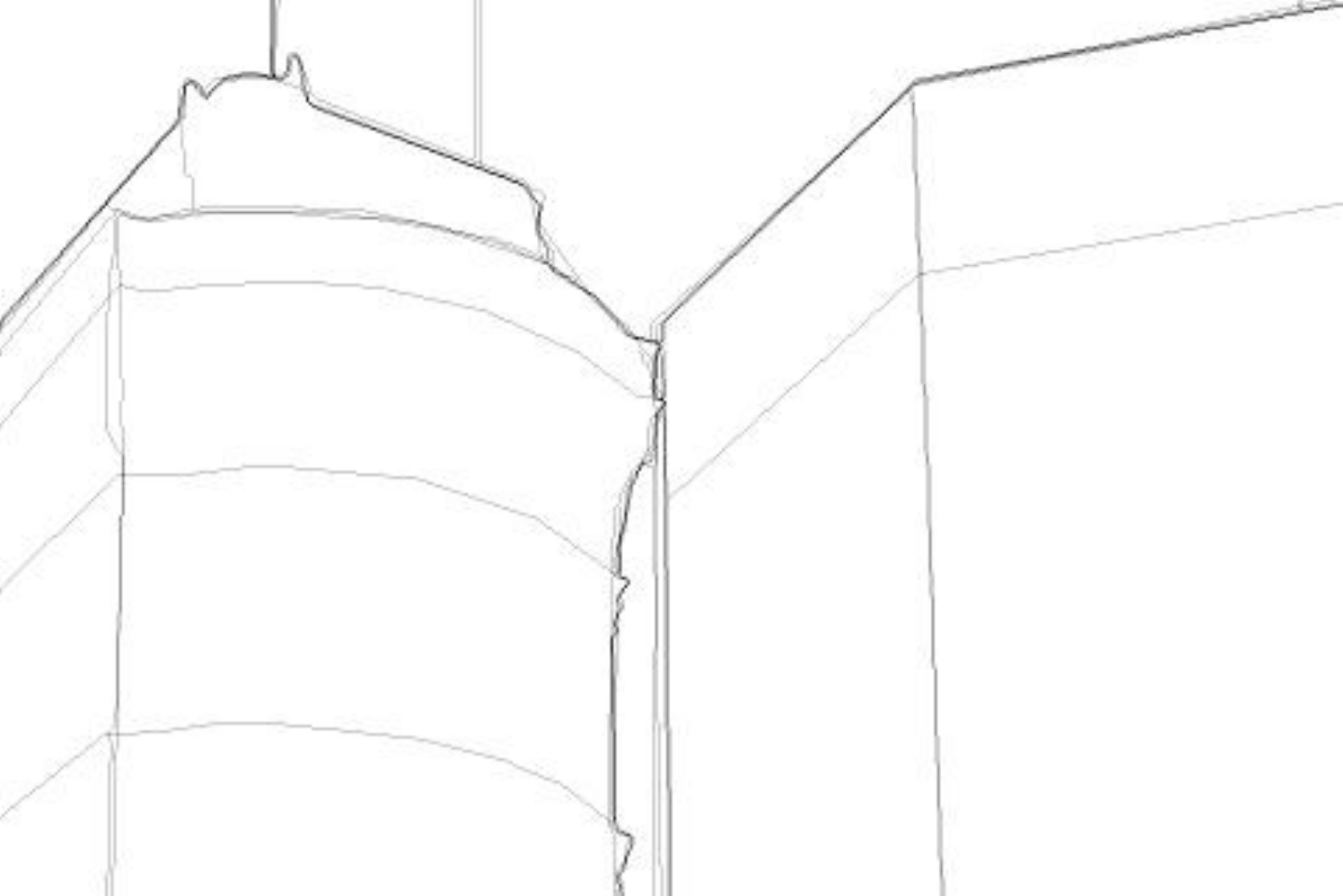} &
\includegraphics[width=0.25\linewidth]{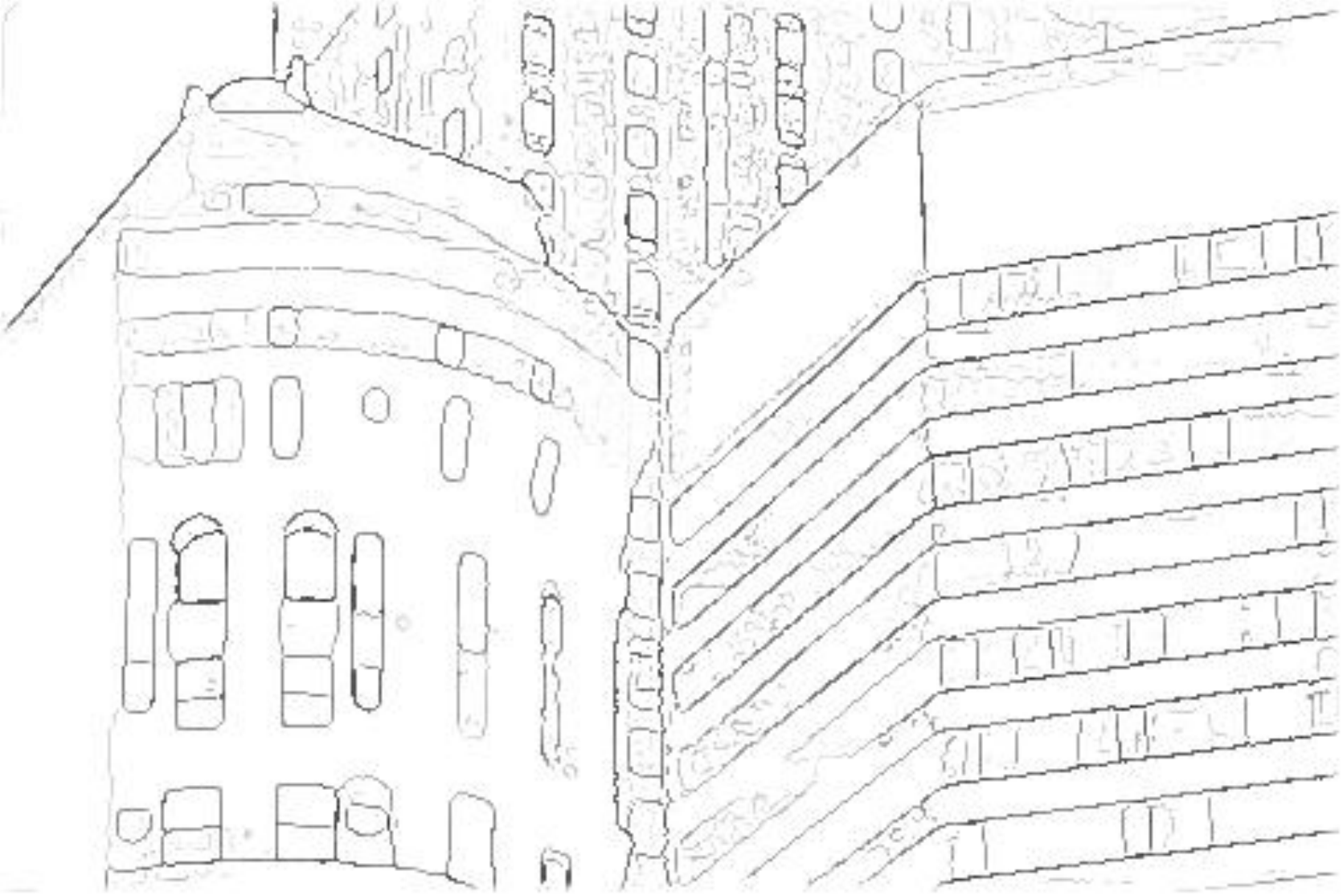} &
\includegraphics[width=0.25\linewidth]{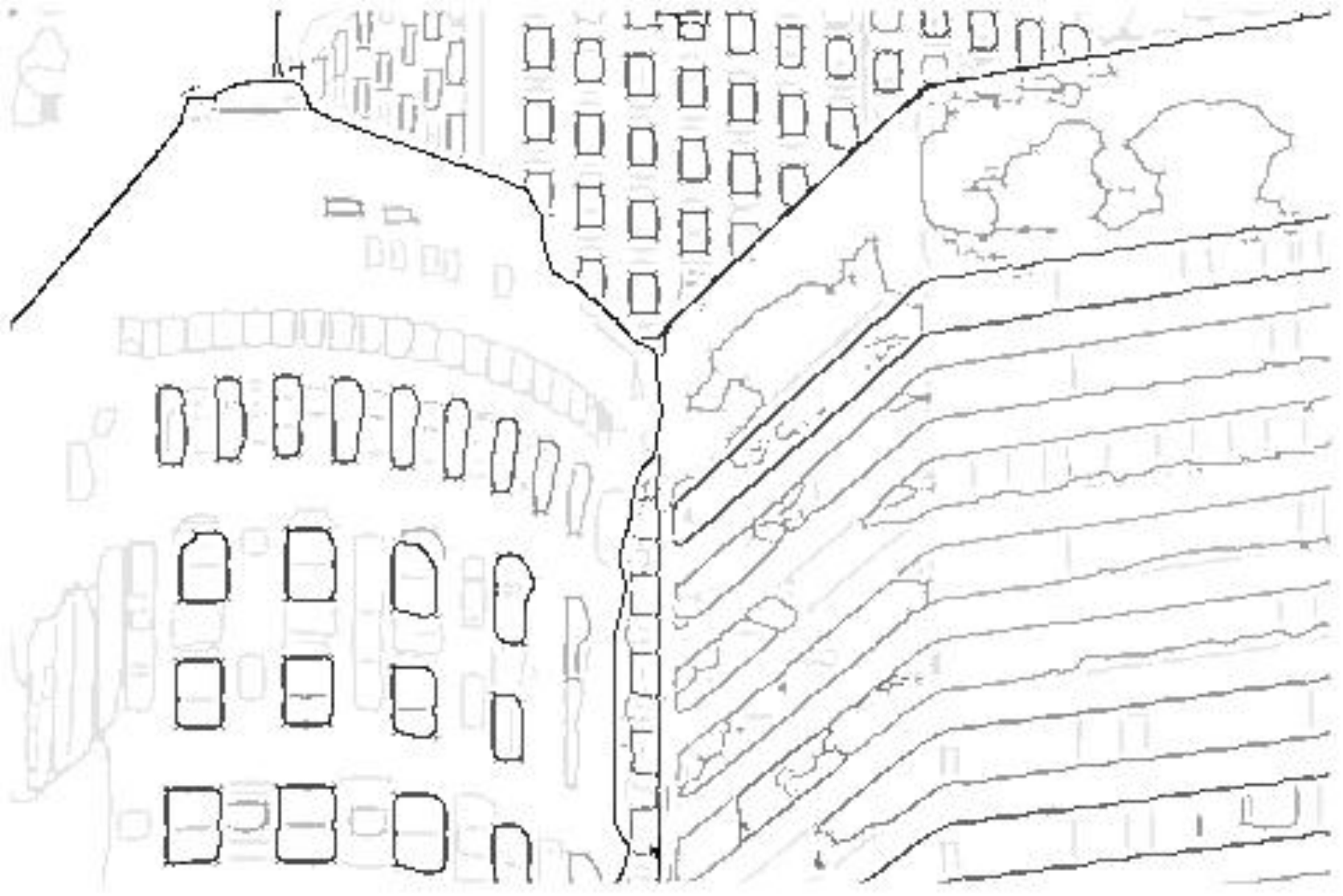} \\
\includegraphics[width=0.25\linewidth]{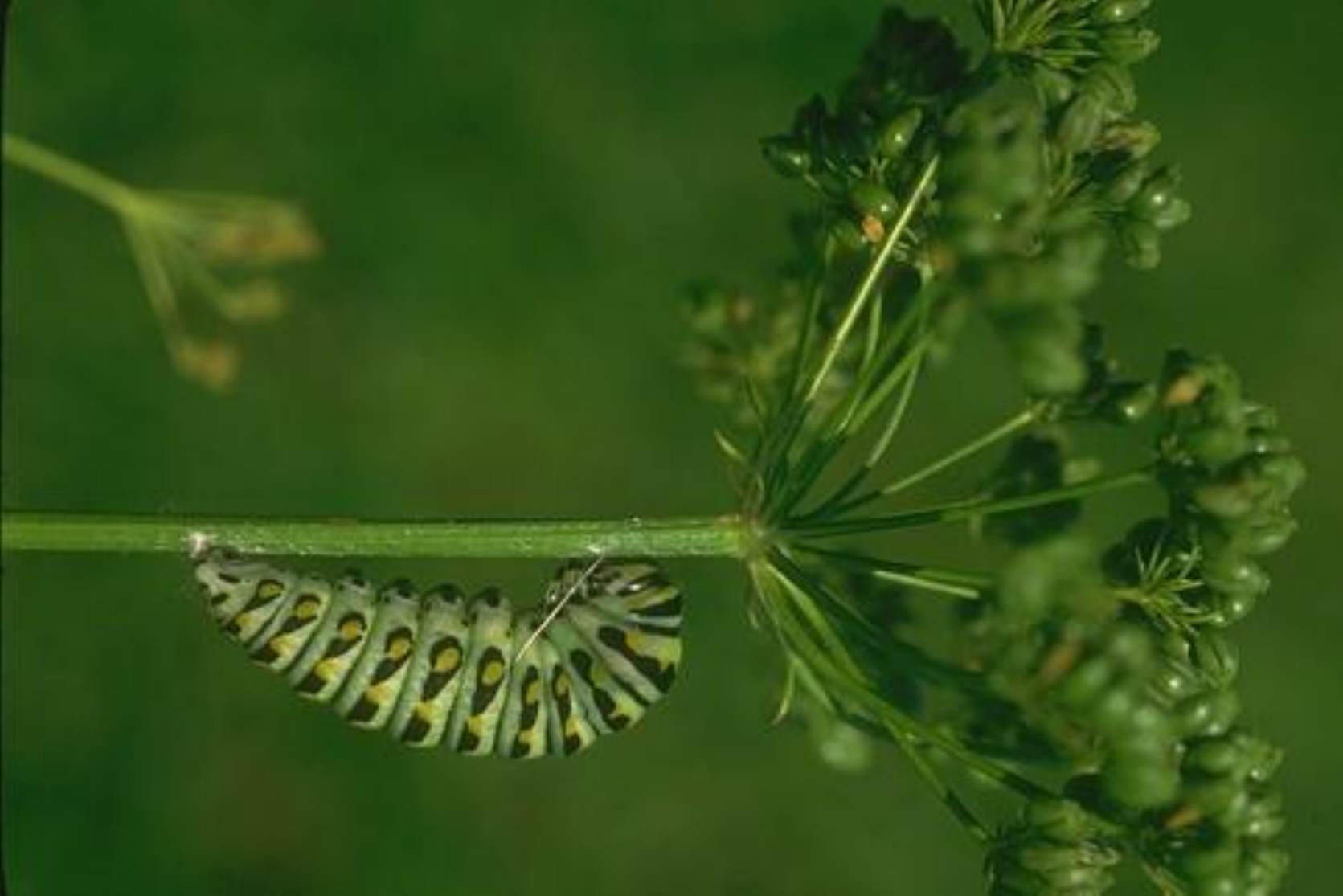} &
\includegraphics[width=0.25\linewidth]{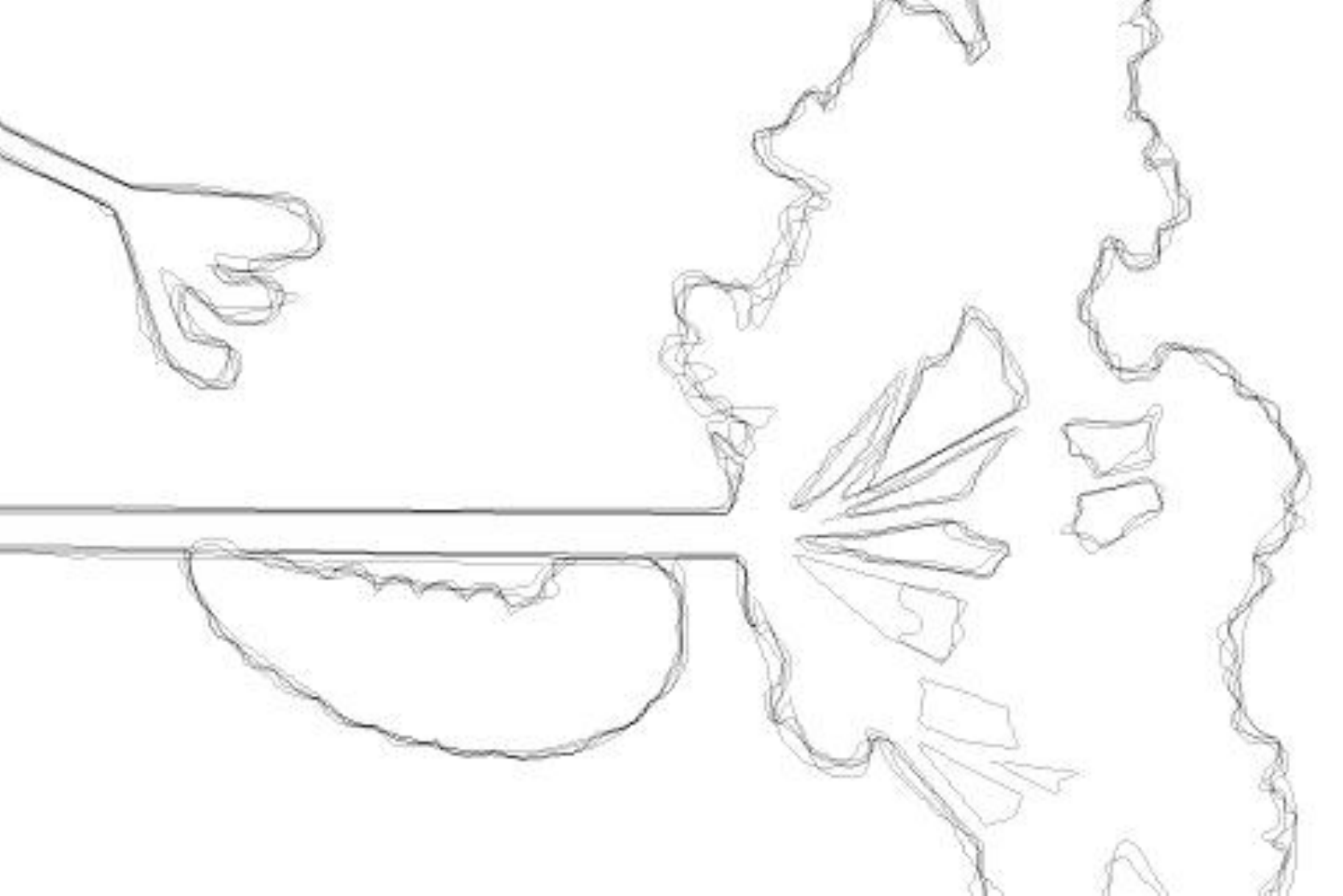} &
\includegraphics[width=0.25\linewidth]{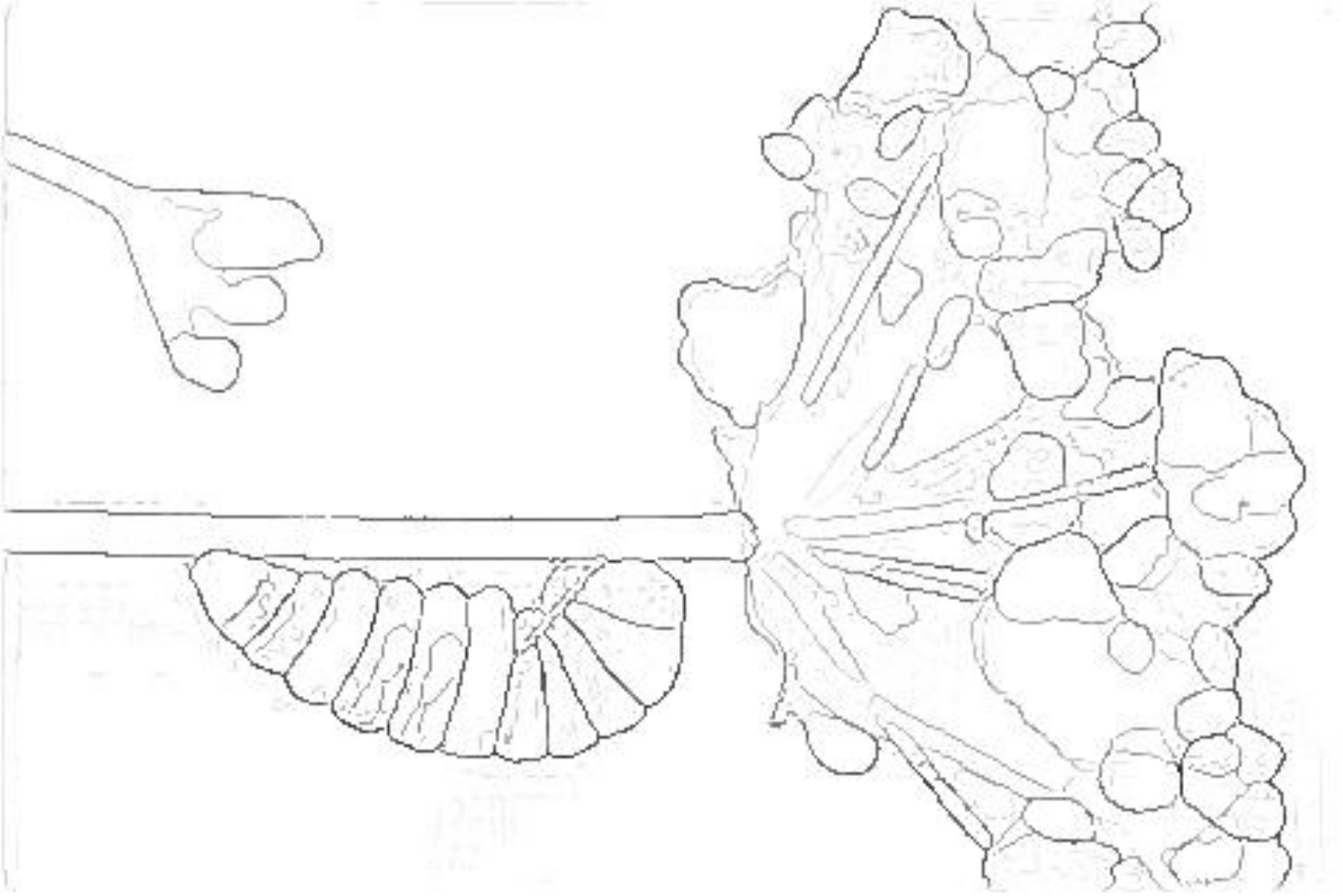} &
\includegraphics[width=0.25\linewidth]{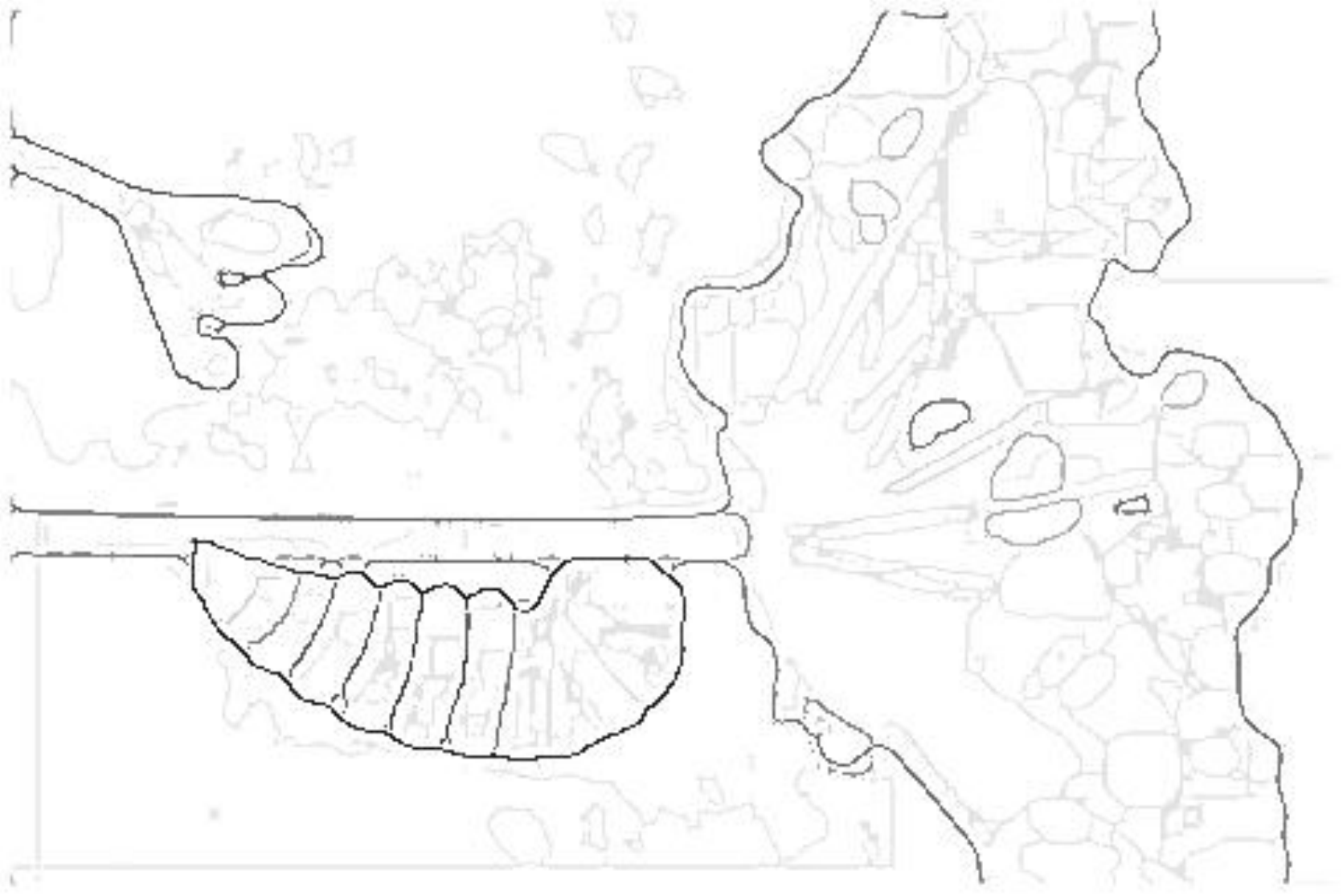} \\
\end{tabular}\vspace{-3mm}
\caption{Zero-shot edge prediction on BSDS500. FastSAM achieves comparable results to SAM.}
\label{fig:edges}
\end{figure*}

\subsection{Zero-Shot Edge Detection}\label{sec:eval:edge}

\paragraph{Approach.} FastSAM is assessed on the basic low-level task of edge detection using BSDS500~\cite{martin2001database,arbelaez2010contour}. Specifically, we select the mask probability map from the results of FastSAM's all-instance segmentation stage. After that, Sobel filtering \cite{sobel19683x3} is applied to all mask probability maps to generate edge maps. Finally, we conclude with the edge NMS~\cite{canny1986computational} step.

\begin{table}[t]
\centering
\tablestyle{9pt}{1.1}
\footnotesize
\begin{tabular}{lc|p{18pt}p{18pt}p{18pt}p{18pt}}

\toprule
method & year & ODS & OIS & AP & R50  \\
\hline
HED~\cite{xie2015holistically} & 2015& .788 & .808 & .840 & .923 \\
EDETR~\cite{pu2022edter} & 2022 & .840 & .858 & .896 & .930 \\
\multicolumn{6}{@{}l}{\emph{zero-shot transfer methods:}} \\
Sobel filter & 1968 & .539 & - & -& - \\
Canny~\cite{canny1986computational} & 1986 & .600 & .640 & .580 & - \\
Felz-Hutt~\cite{felzenszwalb2004efficient} & 2004 & .610 & .640 & .560 & - \\
\sam~\cite{kirillov2023segment} & 2023 &.768 &.786 & .794 & .928 \\
FastSAM & 2023 & .750 & .790 & .793 &.903 \\
\bottomrule
\end{tabular}\vspace{-2mm}
\caption{Zero-shot transfer to edge detection on BSDS500. Evaluation Data of other methods is from \cite{SAM}.}
\label{tab:edges}
\end{table}

\paragraph{Results.} The representative edge maps are illustrated in Figure~\ref{fig:edges}. Upon qualitative observation, it becomes evident that despite FastSAM's significantly fewer parameters (only  68M), it produces a generally good edge map. 
In comparison to the ground truth, both FastSAM and SAM tend to predict a larger number of edges, including some logical ones that aren't annotated in the BSDS500. This bias is quantitatively reflected in Table~\ref{tab:edges}. Table \ref{tab:edges} shows that we achieve similar performance with SAM, specifically a higher R50 and a lower AP.
\footnotetext{\url{https://github.com/facebookresearch/segment-anything}}

\subsection{Zero-Shot Object Proposal Generation}\label{sec:eval:objpro}
 
\paragraph{Background.} Object proposal generation has long been a basic pre-processing step for many computer vision tasks, including general object detection, few-shot object detection, and image understanding. Many famous proposal generation methods witness the evolution of visual recognition in the past decades, as a role of the basic step of visual recognition methods. These proposal generation methods includes EdgeBox\cite{zitnick2014edge}, Geodesic\cite{krahenbuhl2014geodesic}, Selective Search \cite{uijlings2013selective}, MCG \cite{arbelaez2014multiscale}. These years, many deep learning-based methods are proposed like DeepMask \cite{o2015learning}, OLN\cite{kim2022learning}. For example, RCNN-series object detection methods \cite{girshick2015region,girshick2015fast} adopts the  Seletive Search method, and the recently proposed open world detector, UniDetector \cite{wang2023detecting}, adopts the OLN method. Though RPN \cite{ren2015faster} is used by most existing object detectors, it can only generate object proposals of the learned categories, limiting its application in open-vocabulary recognition tasks. Therefore, zero-shot object proposal generation is rather important. A good proposed method is important for the good performance of these visual recognition tasks.

We directly use the generated bounding boxes of the first stage of FastSAM as the object proposals. 
To evaluate the performance, we test on LVIS \cite{gupta2019lvis} and COCO \cite{lin2014microsoft} dataset, following the existing evaluating strategies. Besides this, following the experimental settings of SAM, we also test the mask proposal accuracy by using the all-instance masks of the first stage. 

\paragraph{Details.} We report the results of SAM, ViTDet \cite{li2022exploring}, and our FastSAM on the LVIS dataset. As SAM does not publicize its detailed evaluation codes, we reproduced the category-agnostic mask and box recall using the official LVIS evaluation code \cite{gupta2019lvis}. However, we fail to reproduce the Mask AR results of ViTDet and SAM presented in the SAM's paper \cite{SAM}. Nevertheless, we think our evaluation results still reflect several features of FastSAM compared with SAM.  

\begin{figure}
    \centering
    \includegraphics[width=\columnwidth]{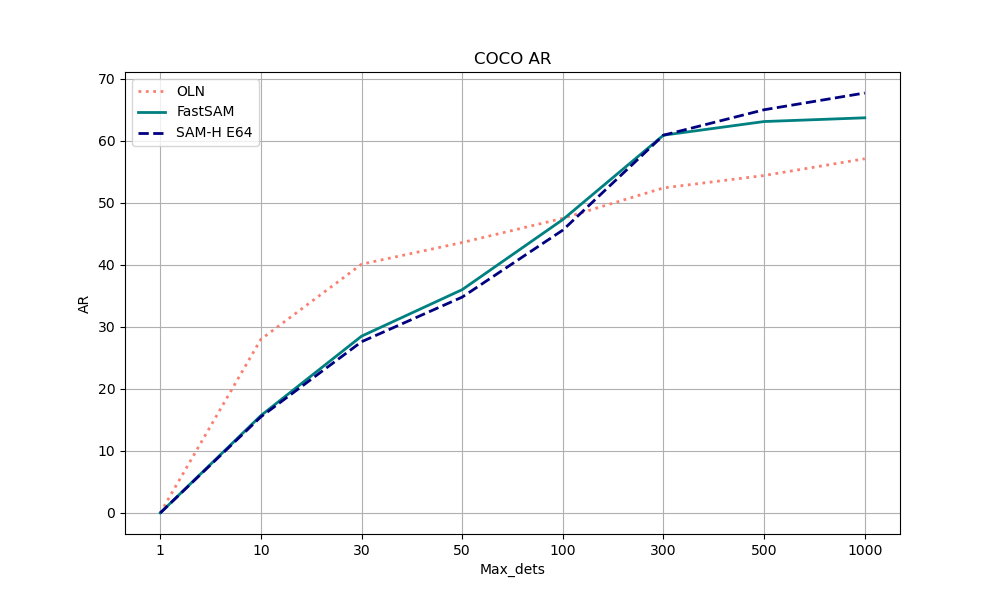}
    \caption{Comparison with OLN \cite{kim2022learning} and SAM-H\cite{SAM}. We test
object proposal methods on all 80 COCO classes}
    \label{fig:cocoar}
\end{figure}

\paragraph{Results.} The results is shown in Table \ref{tab:bbox_proposals_coco}, \ref{tab:bbox_proposals}, and \ref{tab:mask_proposals}. The results show that our method has a significant advantage on the box proposal generation tasks. 
Table \ref{tab:bbox_proposals_coco} presents the Average Recall (AR) of various methods on the COCO validation set. Among these, EdgeBoxes~\cite{zitnick2014edge}, Geodesic~\cite{krahenbuhl2014geodesic}, Sel.Search~\cite{uijlings2013selective}, and MCG~\cite{arbelaez2014multiscale} are methods that do not require training, whereas DeepMask~\cite{o2015learning} and OLN ~\cite{kim2022learning}  are supervised methods that are trained on VOC categories within the COCO training set, and then tested across all categories. In contrast, our approach and SAM~\cite{SAM} implement a fully zero-shot transfer.
From the table, it can be seen that our method and SAM ~\cite{SAM} do not perform as well in AR@10 precision compared to previous supervised methods such as OLN ~\cite{kim2022learning}. However, in AR@1000, our method significantly outperforms OLN ~\cite{kim2022learning}. The reason for this is that previous methods were trained on certain categories in COCO, leading to a higher confidence level in these categories during inference. However, since our method and SAM are zero-shot, this results in balanced confidence levels across different categories, thereby recalling more categories not present in COCO. More comparisons can be seen in Figure ~\ref{fig:cocoar}.

In Table \ref{tab:bbox_proposals}, we report the bbox AR@1000 results of VitDet-H ~\cite{li2022exploring}, SAM ~\cite{SAM}, and our method on the LVIS v1 dataset. Our method substantially surpasses the most computationally intensive model of SAM, SAM-H E64, by over 5\%. However, it falls short compared to VitDet-H~\cite{li2022exploring}, which was trained on the LVIS dataset.
The reason for these results is that during our training process, we used the ground truth (gt) bounding box (bbox) information as a supervisory signal. SAM~\cite{SAM}, on the other hand, only used the mask as a supervisory signal, and its bbox at inference is generated by extracting the outer box from the mask.

From Table \ref{tab:mask_proposals}, our mask proposal generation is relatively lower on Recall. We find this mainly results from that our segmentation mask on small-sized objects is not fine-grained enough. We give more detailed discussions in Section \ref{sec:disc}.

\begin{table}[t]
\centering
 {
 \small
\begin{tabular}{l|cccc}
\hline
 & {AR10} & {AR100} &{AR1000} & {AUC} \\
\hline  
EdgeBoxes~\cite{zitnick2014edge}        & 7.4   & 17.8  & 33.8  & 13.9 \\
Geodesic~\cite{krahenbuhl2014geodesic}  & 4.0   & 18.0  & 35.9  & 12.6 \\
Sel.Search~\cite{uijlings2013selective} & 5.2   & 16.3  & 35.7  & 12.6 \\
MCG~\cite{arbelaez2014multiscale}       & 10.1  & 24.6  & 39.8  & 18.0 \\
\hline
DeepMask~\cite{o2015learning}  & 13.9  & 28.6  & 43.1  & 21.7 \\
 
OLN-Box ~\cite{kim2022learning} & 27.7  & 46.1  & 55.7  & 34.3 \\

\hline
SAM-H E64  & 15.5  & 45.6  & 67.7  & 32.1 \\
SAM-H E32  & 18.5  & 49.5  & 62.5  & 33.7 \\
SAM-B E32  & 11.4  & 39.6  & 59.1  & 27.3 \\
FastSAM (Ours) & 15.7  & 47.3  & 63.7 & 32.2 \\

\hline
\end{tabular}}
\caption{\small{Comparison with learning-free methods on All Categories of COCO}. We report average recall (AR) and AUC of learning free methods, deep learning methods (trained on VOC), and ours vs SAM on {All} generalization. The scores of competing methods are taken from~\cite{kim2022learning}, which test object proposal methods on all 80 COCO classes.}
\label{tab:bbox_proposals_coco}
\end{table}

\begin{table}[t]
\centering
\footnotesize
\begin{tabular}{lc|ccc}
\toprule
\multirow{2}{*}{} & \multicolumn{4}{c}{bbox AR@1000}\\

method & all & small & med. & large \\
\hline
ViTDet-H~\cite{li2022exploring} & 65.0 & 53.2 & 83.3 & 91.2 \\
\multicolumn{4}{@{}l}{\emph{zero-shot transfer methods:}} \\
\sam-H E64 & 52.1 & 36.6 & 75.1 & 88.2  \\
\sam-H E32 & 50.3 & 33.1 & 76.2 & 89.8  \\
\sam-B E32 & 45.0 & 29.3 & 68.7 & 80.6  \\
FastSAM (Ours) & 57.1 & 44.3 & 77.1 & 85.3  \\
\bottomrule
\end{tabular}
\vspace{-2mm}
\caption{Object proposal generation on LVIS v1. FastSAM and SAM is applied zero-shot, \ie it was not trained for object proposal generation nor did it access LVIS images or annotations.}
\label{tab:bbox_proposals}
\end{table}

  \definecolor{gray1}{RGB}{169,169,169}
\begin{table}[t]
\centering
\tablestyle{2.8pt}{1.1}
\footnotesize 
\begin{tabular}{l|p{18pt}p{18pt}|p{18pt}p{18pt}p{18pt}p{18pt}p{12pt}}
\toprule
\multirow{2}{*}{} & \multicolumn{7}{c}{mask AR@1000}\\
method & all & small & med. & large & freq. & com. & rare \\
\hline
\multicolumn{8}{@{}l}{\emph{\textcolor{gray1}{results reported in the SAM paper:}}} \\

\textcolor{gray1}{ViTDet-H}~\cite{li2022exploring} & \textcolor{gray1}{63.0} & \textcolor{gray1}{51.7} & \textcolor{gray1}{80.8} & \textcolor{gray1}{87.0} & \textcolor{gray1}{63.1} & \textcolor{gray1}{63.3} & \textcolor{gray1}{58.3} \\

\textcolor{gray1}{\sam ~\cite{SAM} -- single out.} & \textcolor{gray1}{54.9} & \textcolor{gray1}{42.8} & \textcolor{gray1}{76.7} & \textcolor{gray1}{74.4} & \textcolor{gray1}{54.7} & \textcolor{gray1}{59.8} & \textcolor{gray1}{62.0} \\
\textcolor{gray1}{\sam ~\cite{SAM}} & \textcolor{gray1}{59.3} & \textcolor{gray1}{45.5} & \textcolor{gray1}{81.6} & \textcolor{gray1}{86.9} & \textcolor{gray1}{59.1} & \textcolor{gray1}{63.9} & \textcolor{gray1}{65.8} \\

\hline
\multicolumn{8}{@{}l}{\emph{results after our replication:}} \\

ViTDet-H~\cite{li2022exploring} & 59.9 & 48.3 & 78.1 & 84.8 & - & - & - \\
\sam-H E64 & 54.2 & 39.6 & 77.9 & 83.9  \\
\sam-H E32 & 51.8 & 35.2 & 78.7 & 85.2 & - & - & - \\
\sam-B E32 & 45.8 & 31.1 & 70.5 & 73.6 & - & - & - \\
FastSAM (Ours) & 49.7 & 35.6 & 72.7 & 77.6 & - & - & - \\
\bottomrule

\end{tabular}
\vspace{-2mm}
\caption{Object proposal generation on LVIS v1. FastSAM and SAM are applied zero-shot, \ie it was not trained for object proposal generation nor did it access LVIS images or annotations.}
\label{tab:mask_proposals}
\end{table}

\begin{figure}[t]
\vspace{-0.3cm}
	\centering
	\includegraphics[width=1\linewidth]{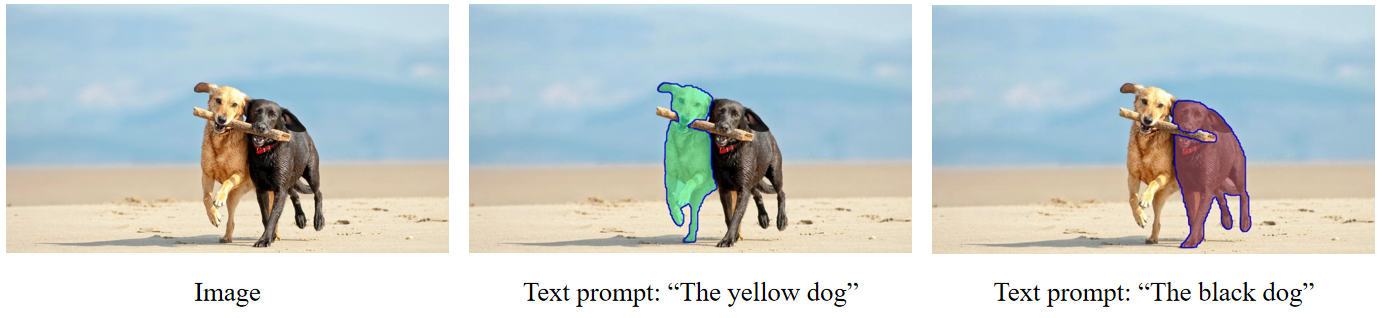}
	\vspace{-0.1cm}
	\caption{Segmentation results with text prompts}
	\vspace{-0.1cm}
	\label{fig:textprompts}
\end{figure}
\begin{figure*}[t]
\vspace{-0.3cm}
	\centering
	\includegraphics[width=0.9\linewidth]{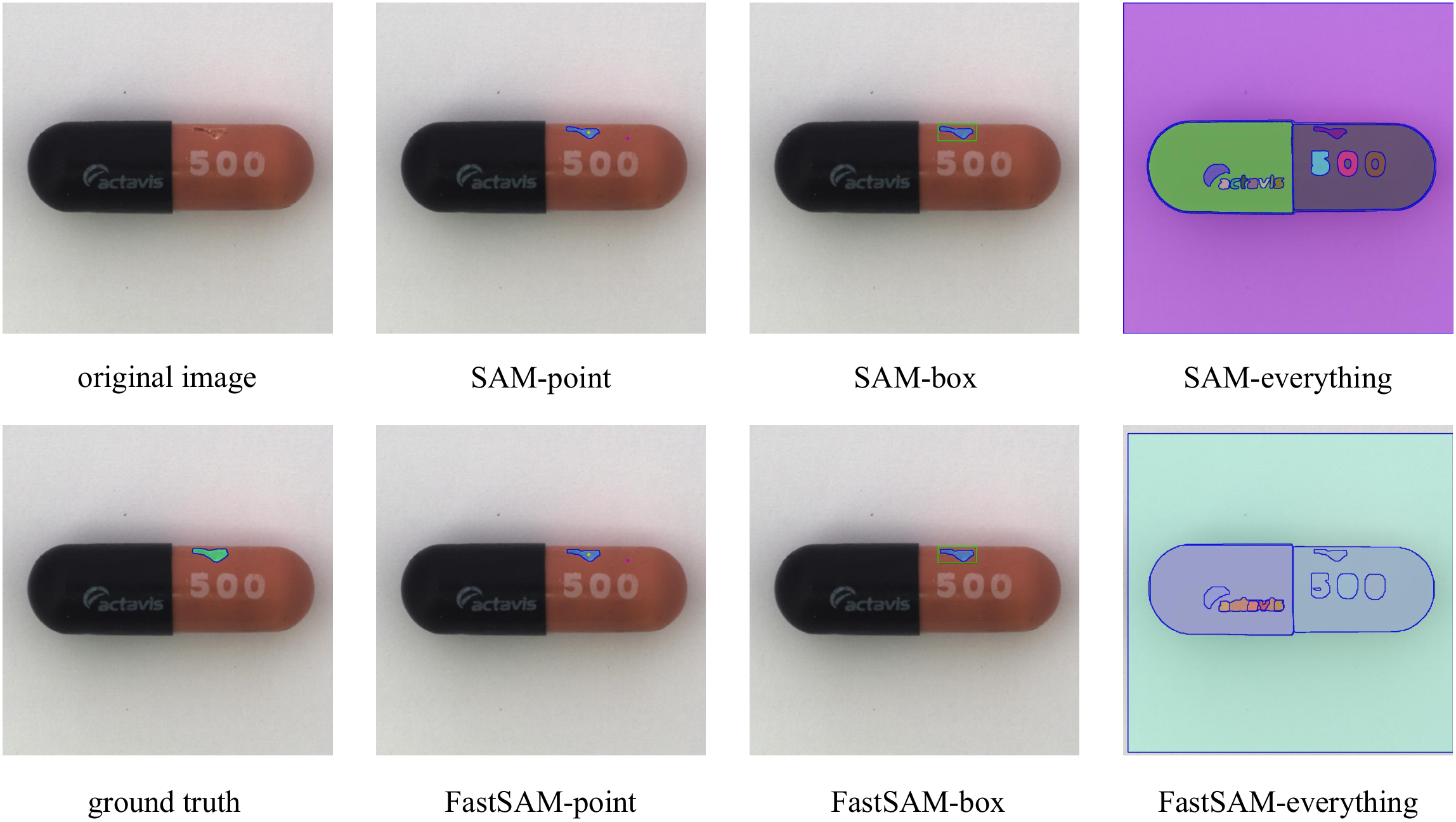}
	\vspace{-0.1cm}
	\caption{Application on \textbf{anomaly detection}, where SAM-point/box/everything means using point-prompt, box-prompt, and everything modes respectively.}
	\vspace{-0.1cm}
	\label{fig:ANO}
\end{figure*}

\subsection{Zero-Shot Instance Segmentation}\label{sec:eval:instseg}

\paragraph{Approach.} Similarly to the SAM method, we accomplish the instance segmentation task by utilizing the bounding box (bbox) generated by ViTDet \cite{li2022exploring} as the prompt. As described by Section \ref{sec:method:PGS}, we choose the mask with the highest Intersection over Union (IoU) with the bbox as the predicted mask.

\paragraph{Results.} Table \ref{tab:instance_segmentation} gives the evaluation results. On this task, we fail to achieve a high AP. We infer that this mainly because of the segmentation mask accuracy or the box-based mask selection strategy.  Section \ref{sec:disc} gives several examples.

\begin{table}[t]
\centering
\tablestyle{2.2pt}{1.1}
\footnotesize
\begin{tabular}{@{}lp{18pt}p{18pt}p{18pt}p{18pt}|p{18pt}p{18pt}p{18pt}p{18pt}}
\toprule
 \multirow{2}{*}{} & \multicolumn{4}{c}{COCO~\cite{Lin2014}} & \multicolumn{4}{c}{LVIS v1~\cite{gupta2019lvis}}\\
method & AP & AP\textsuperscript{S} & AP\textsuperscript{M} & AP\textsuperscript{L} & AP & AP\textsuperscript{S} & AP\textsuperscript{M} & AP\textsuperscript{L} \\
\hline
ViTDet-H~\cite{li2022exploring} & 51.0 & 32.0 & 54.3 & 68.9 & 46.6 & 35.0 & 58.0 & 66.3\\
\multicolumn{9}{@{}l}{\emph{zero-shot transfer methods (segmentation module only):}} \\
\sam 
& 46.5 & 30.8 & 51.0 & 61.7 & 44.7 & 32.5 & 57.6 & 65.5\\
FastSAM 
& 37.9 & 23.9 & 43.4 & 50.0 & 34.5 & 24.6 & 46.2 & 50.8 \\
\bottomrule
\end{tabular}
\vspace{-2mm}
\caption{Instance segmentation results. Fastsam is prompted with ViTDet boxes to do zero-shot segmentation. The fully-supervised ViTDet outperforms \sam, but the gap shrinks on the higher-quality LVIS masks.}
\label{tab:instance_segmentation}\vspace{-1mm}
\end{table}

\subsection{Zero-Shot Object Localization with Text Prompts }\label{sec:eval:text2mask}

\paragraph{Approach.} Finally, we consider an even high-level task, i.e. segmenting objects by free-form texts. This experiment is to show the FastSAM ability of processing text prompts like SAM. Different wit SAM, FastSAM doesn't need to modify the training procedure. It directly runs text through CLIP’s text encoder and then uses the resulting text embedding to find the most similar mask at inference time.

\begin{figure*}[t]
\vspace{-0.3cm}
	\centering
	\includegraphics[width=0.9\linewidth]{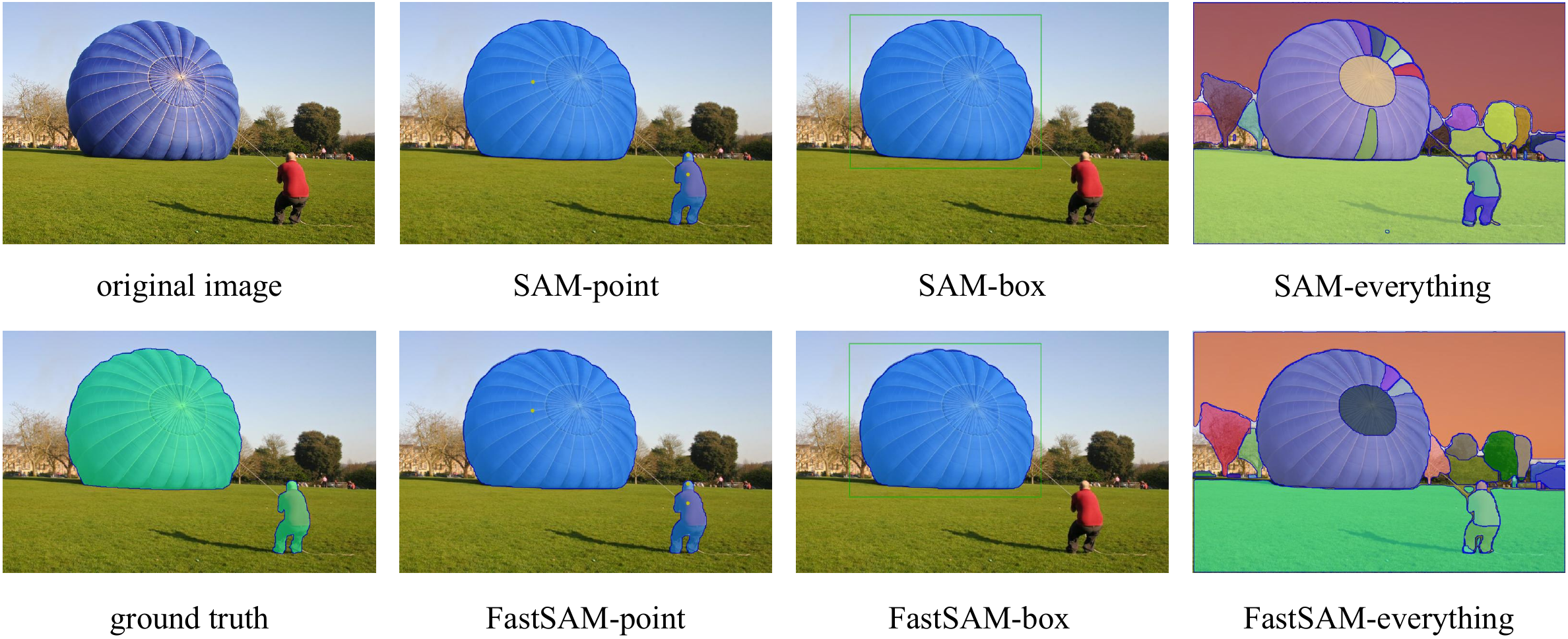}
	\vspace{-0.1cm}
	\caption{Application on \textbf{salient object segmentation}, where SAM-point/box/everything mean using point-prompt, box-prompt, and everything modes respectively.}
	\vspace{-0.1cm}
	\label{fig:SOD}
\end{figure*}

\begin{figure*}
        \vspace{-0.3cm}
	\centering
	\includegraphics[width=0.9\linewidth]{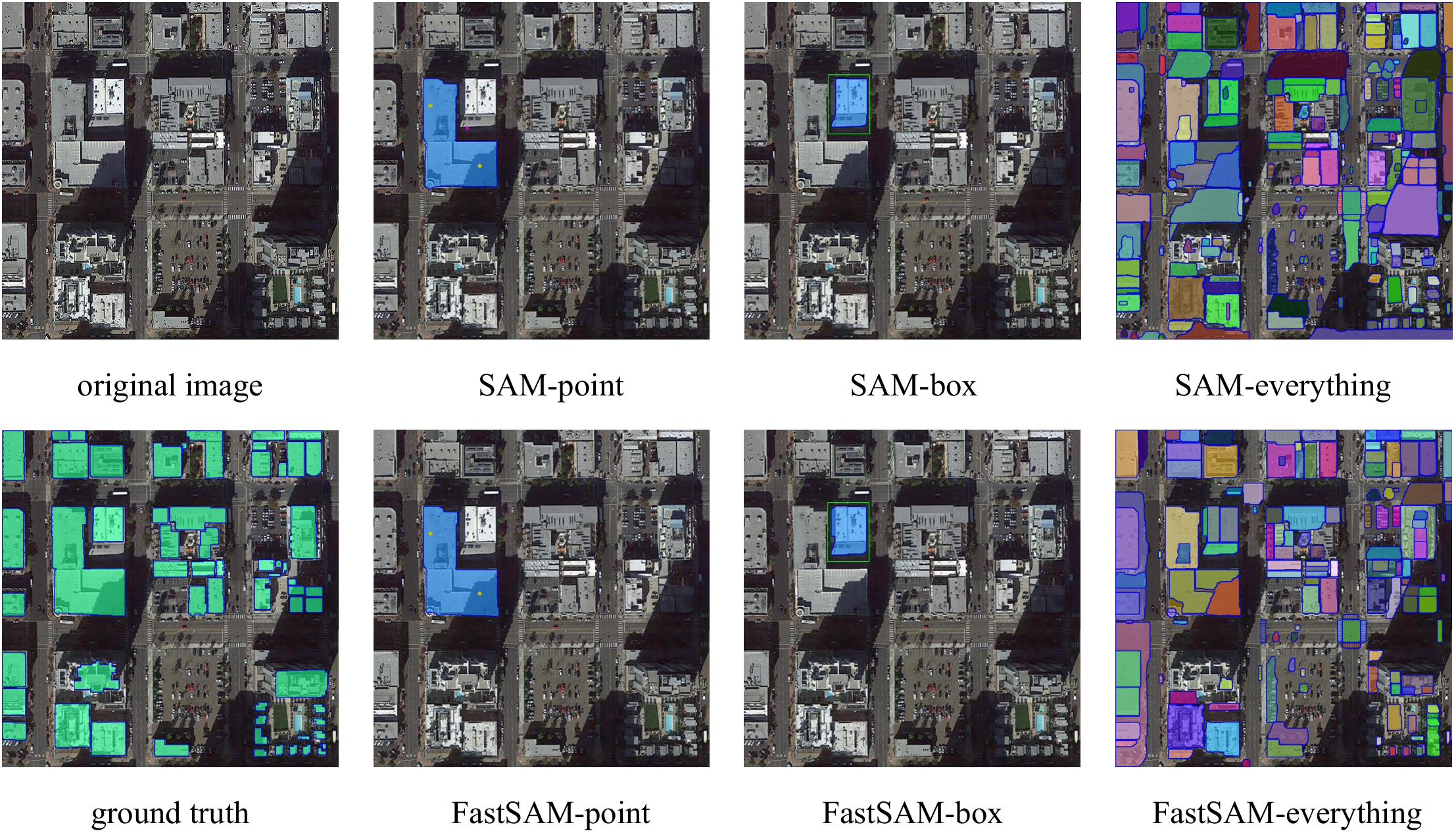}
	\vspace{-0.1cm}
	\caption{Application on \textbf{building extracting}, where SAM-point/box/everything means using point-prompt, box-prompt, and everything modes respectively.}
	\vspace{-0.1cm}
	\label{fig:build}
\end{figure*}

\paragraph{Results.} We show qualitative results in \fig{fig:textprompts}. FastSAM can segment objects well based on the text prompts. Nevertheless, the running speed of the text-to-mask segmentation is not satisfying, since each mask region is required to be fed into the CLIP feature extractor. How to combine the CLIP embedding extractor into the  FastSAM's backbone network remains an interesting problem with respect to the model compression.

\section{Real-world Applications}\label{sec:eval:app}

In this section, we evaluate the performance of FastSAM across different application scenarios and analyze its advantages and limitations. We showcase visualizations of FastSAM's segmentation using point-prompt, box-prompt, and everything modes, and compare it with SAM and ground truths.

\paragraph{Anomaly Detection.}

As detailed in~\cite{MVTec}, anomaly detection is a task that aims to distinguish between defective and normal samples in the manufacture. FastSAM is evaluated using the MVTec AD dataset~\cite{MVTec}, with results displayed in \fig{fig:ANO}. Under everything mode, FastSAM can segment nearly all regions similar to SAM, but with a lower level of precision compared to SAM. In addition, the mask for the background does not completely cover the entire background, which is an inherent characteristic of YOLACT~\cite{bolya2019yolact}. By foreground/background points (yellow and magenta points in FastSAM-point respectively) or box-guided selection, FastSAM can segment on the exact defective regions.

 \begin{figure*}[t]
\vspace{-0.3cm}
	\centering
	\includegraphics[width=0.79\linewidth]{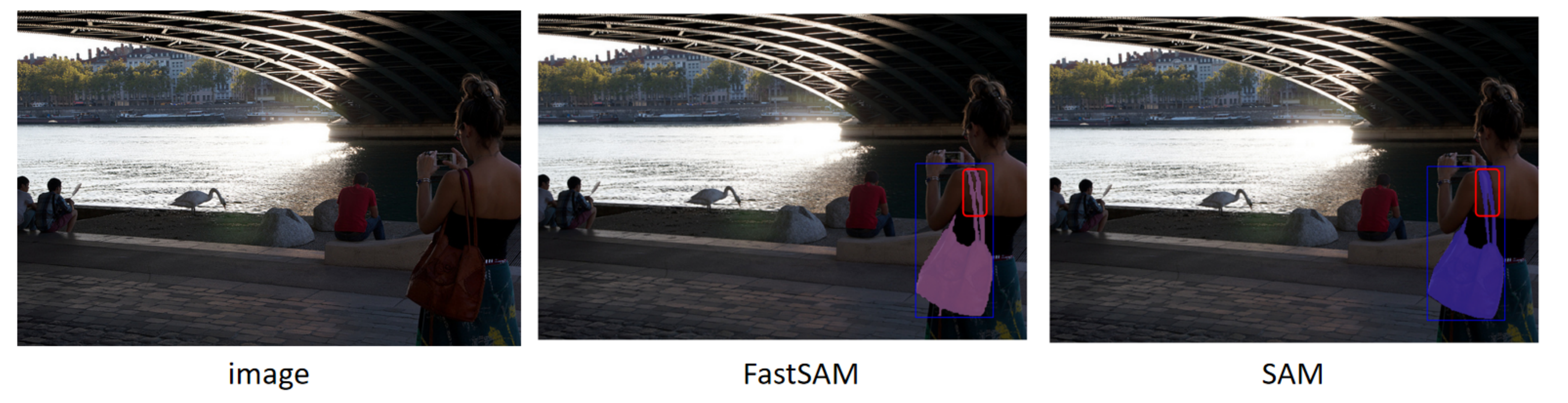}
	\vspace{-0.1cm}
	\caption{FastSAM generates finer segmentation masks on the narrow region of the large objects}
	\vspace{-0.1cm}
	\label{fig:discussion}
\end{figure*}

 \begin{figure*}[t]
\vspace{-0.1cm}
	\centering
	\includegraphics[width=0.79\linewidth]{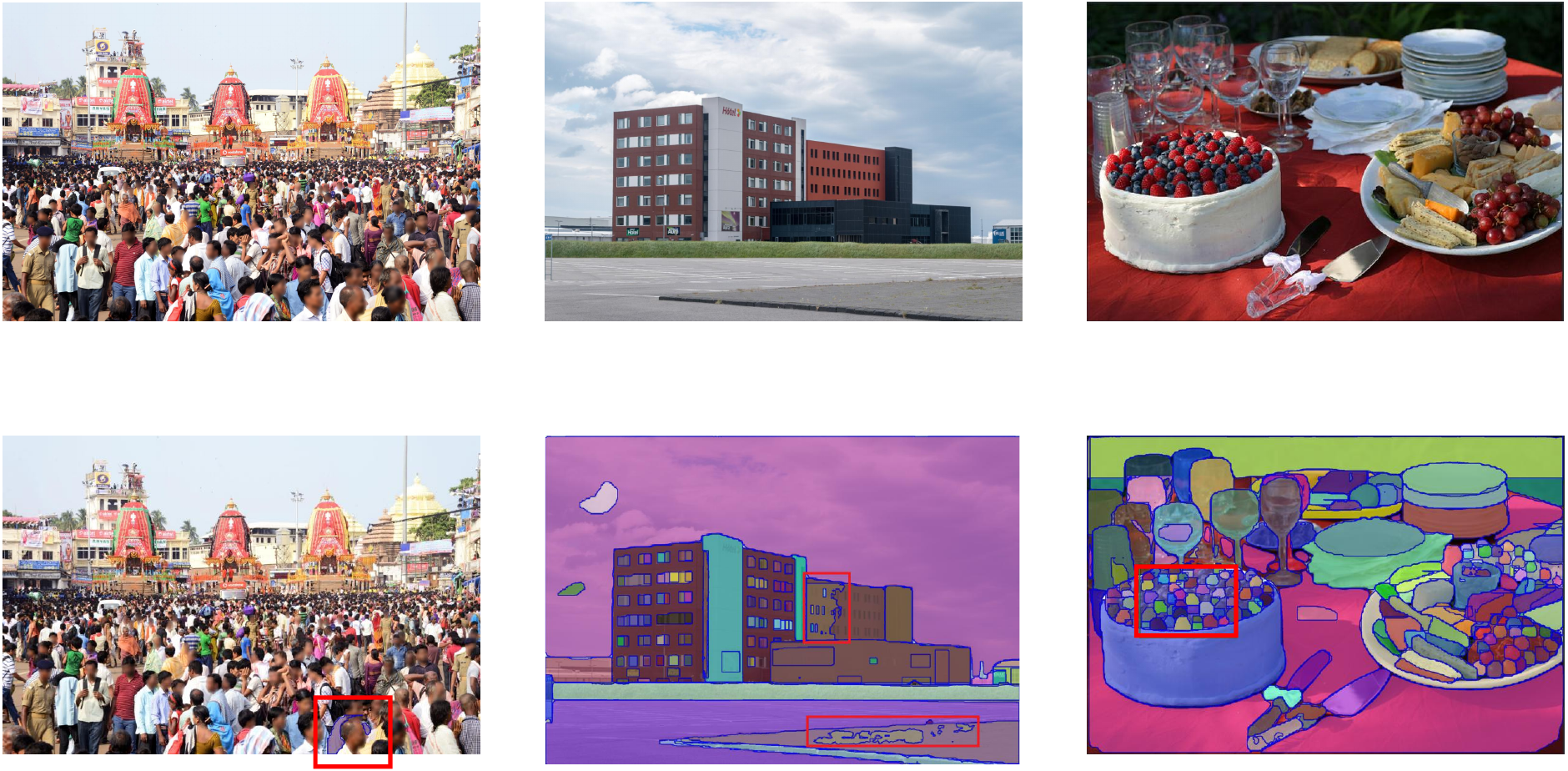}
	\vspace{-0.1cm}
	\caption{Some examples for the bad case of FastSAM.}
	\vspace{-0.1cm}
	\label{fig:weakness}
\end{figure*}

\paragraph{Salient Object Segmentation.}

The aim of salient object segmentation~\cite{borji2019salient1} is to segment the most attention-grabbing objects from an image. This task is class-agnostic, setting it apart from semantic segmentation. We apply FastSAM to the well-known saliency dataset, ReDWeb-S~\cite{WebS}. As presented in \fig{fig:SOD}, FastSAM exhibited only a minor difference from SAM under everything mode, as it segment fewer background objects which are irrelevant to the task. By points-guided selection, such as yellow points in FastSAM-point, we can obtain masks of all objects of interest. The segmentation result of FastSAM-point is nearly identical to that of the SAM-point and the ground truth, with only minor details lost at the edges. The object of interest can also be selected by box prompt, such as the green box in FastSAM-box. However, it is impossible to select multiple objects with a single box, which even SAM-box cannot realize.

\paragraph{Building Extracting.}

Building Extracting from optical remote sensing imagery has a wide range of applications, such as urban planning. We evaluate FastSAM on the dataset proposed by~\cite{ji2018fully}. As demonstrated in Fig.\ref{fig:build}, FastSAM performs well in segmenting regularly shaped objects, but segments fewer regions related to shadows compared to SAM. We can also select regions of interest with point-prompt and box-prompt, as presented in FastSAM-point and FastSAM-box. It is worth noting that we position a point in a shadow region in FastSAM-point. However, the correct mask for the building can still be obtained by merging based on this point. This indicates that our method can resist the interference of noise to some extent.

\section{Discussion}\label{sec:disc}
Generally, the proposed FastSAM achieves a comparable performance with SAM and runs 50x faster than SAM (32$\times$32) and 170x faster than SAM (64$\times$64). The running speed makes it a good choice for industrial applications, such as road obstacle detection, video instance tracking, and image manipulation. On some images, FastSAM even generates better masks for large objects, as shown in Figure \ref{fig:discussion}.

\paragraph{Weakness.} However, as presented in the experiments, our box generation has a significant advantage, but our mask generation performance is below SAM. We visualize these examples in Figure \ref{fig:weakness}. We find that FastSAM has the following features. 
\begin{itemize}
    \item The low-quality small-sized segmentation masks have large confidence scores. We think this is because the confidence score is defined as the bbox score of YOLOv8, which is not strongly related to the mask quality. Modifying the network to predict the mask IoU or other quality indicators is a way to improve that.
    
    \item The masks of some of the tiny-sized objects tend to be near the square. Besides, the mask of large objects may have some artifacts on the border of the bounding boxes. This is the weakness of the YOLACT method. By enhancing the capacity of the mask prototypes or reformulating the mask generator, the problem is expected to solve.
\end{itemize}

Moreover, since we only use 1/50 of all SA-1B datasets, the model's performance can also be further enhanced by utilizing more training data.

\section{Conclusion}
In this paper, we rethink the segment of anything task and the model architecture choosing, and propose an alternative solution with 50 times faster running speed than SAM-ViT-H (32$\times$32). The experiments show that FastSAM can solve multiple downstream tasks well. Still, there are several weaknesses that can be improved for FastSAM, like the scoring mechanism and the instance mask-generating paradigm. These problems are left for future study.

  {\small
  \bibliographystyle{ieee_fullname}
  \bibliography{egbib}
  }

\end{document}